\documentclass[10pt,twocolumn,letterpaper]{article}

\usepackage{iccv}
\usepackage{times}
\usepackage{epsfig}
\usepackage{graphicx}
\usepackage{amsmath}
\usepackage{amssymb}
\usepackage{subfigure}
\usepackage{multirow}
\usepackage{bm}
\newtheorem{lemma}{Lemma}
\usepackage{booktabs}

\usepackage[pagebackref=true,breaklinks=true,letterpaper=true,colorlinks,bookmarks=false]{hyperref}
\usepackage[capitalize]{cleveref}
\crefname{section}{Sec.}{Secs.}
\Crefname{section}{Section}{Sections}
\Crefname{table}{Table}{Tables}
\crefname{table}{Tab.}{Tabs.}

\iccvfinalcopy 

\def\iccvPaperID{7471} 

\ificcvfinal\fi

\begin{document}

\title{RdimKD: Generic Distillation Paradigm by Dimensionality Reduction}
\author{
    Yi Guo,
    Yiqian He,
    Xiaoyang Li,
    Haotong Qin,
    Van Tung Pham,\\
    Yang Zhang,
    Shouda Liu \\
    bytedance\\
    {\fontsize{8.5pt}{12} \texttt{\{guoyi.0,heyiqian.11,lixiaoyang.x,qinhaotong,van.pham,zhangyang.elfin,liushouda\}@bytedance.com}}
}


\maketitle
\ificcvfinal\thispagestyle{empty}\fi

\begin{abstract}
Knowledge Distillation (KD) emerges as one of the most promising compression technologies to run advanced deep neural networks on resource-limited devices. In order to train a small network (student) under the guidance of a large network (teacher), the intuitive method is regularizing the feature maps or logits of the student using the teacher's information. However, existing methods either over-restrict the student to learn all information from the teacher, which lead to some bad local minimum, or use various fancy and elaborate modules to process and align features, which are complex and lack generality. In this work, we proposed an abstract and general paradigm for the KD task, referred to as \textbf{DIM}ensionality \textbf{R}eduction KD (RdimKD), which solely relies on dimensionality reduction, with a very minor modification to naive $\ell^2$ loss.
RdimKD straightforwardly utilizes a projection matrix to project both the teacher's and student's feature maps onto a low-dimensional subspace, which are then optimized during training.
RdimKD achieves the goal in the simplest way that not only does the student get valuable information from the teacher, but it also ensures sufficient flexibility to adapt to the student's low-capacity reality. Our extensive empirical findings indicate the effectiveness of RdimKD across various learning tasks and diverse network architectures. 

\end{abstract}

\section{Introduction}
With the increasing and extensive application of Deep Neural Networks (DNNs) in industry, model compression technologies~\cite{guo2021gdp,krishnamoorthi2018quantizing,lin2022knowledge,zoph2016neural} have been widely studied to deploy deep models on storage and computation limited hardware. Among these technologies, Knowledge Distillation (KD)~\cite{2015Distilling} attracts attention from academia and industry for its high architecture adaptability and compression performance. 

The essence of knowledge distillation lies in how to obtain valuable knowledge from the teacher network. 
For example, soft labels can better reflect the distribution information between categories than hard one-hot labels~\cite{2015Distilling,ba2014deep}. To extend knowledge distillation to more general and complex scenarios, more and more works~\cite{romero2014fitnets,yang2022masked,zagoruyko2016paying,lin2022knowledge} are also exploring the distillation from intermediate feature maps as regularization to assist the training of student networks further. 
A common method, \eg~\cite{romero2014fitnets, jiao2019tinybert,li2017mimicking,yang2021knowledge}, is to use a naive $\ell^2$ loss on the original feature maps of the teacher and student. Specifically,  let $F_t, F_s$ be the feature maps to be distilled of the teacher and student, then the KD loss can be described as
\begin{equation}
    \min \mathcal{L}_{KD} = \min \|F_t-r_{\theta}(F_s)\|^2
    \label{naive l2 loss}
\end{equation}
where $r_{\theta}(\cdot)$ is a learnable transformation layer needed when the shapes of the feature maps mismatch, $\theta$ the learnable parameters, $\| \cdot \|$ the Frobenius norm for matrix.

Intuitively, it is sub-optimum to force the student to get all the information of the teacher in a way like~\cref{naive l2 loss} because of the difference in network capacity, the randomness of initialization, and/or the difficulty of optimization. It can be demonstrated experimentally in~\cref{Experiments} that a simple $\ell^2$ loss between feature maps (with the same shapes) of teacher and student does not bring enough performance improvement for the student.  
So, naturally, we may require the student to only learn some useful information from the teacher, while maintaining a certain degree of flexibility to adapt to the reality of its low capacity. Also, these methods come at a cost of additional modules to be trained as well as more hyper-parameter to be finetuned.

As a result, instead of applying $\ell^2$ loss on the original feature maps, some works~\cite{kim2018paraphrasing,zagoruyko2016paying,zhou2020channel,heo2019knowledge} manipulate and align the feature maps in some fancy and less explainable ways. For example, ~\cite{zagoruyko2016paying} calculates the attention of the feature maps by pooling along the channel dimension, while~\cite{zhou2020channel} performs along the width and height dimensions, and~\cite{yim2017gift} generates FSP matrix from two layers to represent the knowledge flow. Methods in this category are essentially designed to increase students' freedom of learning without over-restricting their flexibility and to only get some valuable information from the teacher network. But these fancy and specific designs are too elaborate to be essential, and we want to reveal the essence of knowledge distillation at a more abstract and higher level.

\begin{figure}[t]
  \centering
   \includegraphics[width=\linewidth]{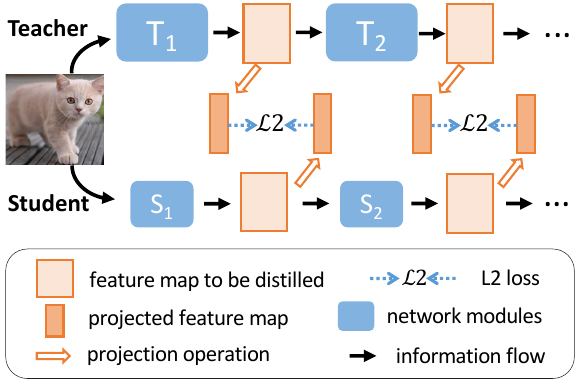}
   \caption{The overall framework of RdimKD. 
   The student's and teacher's feature maps are projected onto a low-dimensional subspace by a matrix, and then the simple $\ell^2$ loss is implemented. The projection of the teacher only allows some valuable knowledge to be transferred out, while the projection of the student leaves the complementary subspace as freedom of the student.}
   \label{fig:network}
\end{figure}

We propose a simple, generic, and effective knowledge distillation method named RdimKD from a more abstract, higher-level, and more essential perspective: {\bf \emph{ dimensionality reduction}}. RdimKD is 
based on dimensionality reduction itself to ensure that students can focus on valuable information from teachers and enjoy enough flexibility. The proposed framework, shown in ~\cref{fig:network}, can be formally described as  
\begin{equation}
\min \mathcal{L}_{KD} = \min \|F_tK-F_sK\|^2
\label{peojection}
\end{equation}
where $K$ is the projection matrix. We do not use the function $r_{\theta}(\cdot)$ with learnable $\theta$ in RdimKD because of two reasons. On the one hand, the function is somehow tricky to design in specific tasks, and we hope to abstract the essence of knowledge distillation in a more generic way; on the other hand, when optimizing~\cref{naive l2 loss}, this function will make the student be lazy to some extend, and it just relies on $\theta$ to reconcile the difference with the teacher instead of getting knowledge from it.

Note that the proposed method requires $F_t$ and $F_s$ to have the same dimension. 
To allow more flexible architecture for teacher and student, there are two solutions to bypass the above requirement: 1. train a teacher with the same shapes at distillation positions; 2. set the distillation position at a linear transformation (e.g. linear or convolution layer) $f: \mathbb{R}^{p}\rightarrow\mathbb{R}^q$ of the student. $f$ can be split into two transformations, $f_1: \mathbb{R}^{p}\rightarrow\mathbb{R}^{t}$ and $f_2: \mathbb{R}^{t}\rightarrow\mathbb{R}^{q}$, where $t$ is the dimension of teacher. In this way, teacher and student have the same dimension $t$ and can be distilled by RdimKD. During inference, $f_1$ and $f_2$ can be merged into $f$ without changing the original structure of the student.

RdimKD focuses on the concept of dimensionality reduction itself, regardless of the specific reduction methods. To show this, we provide three reduction methods that are very common in statistical machine learning, \textit{i.e.}, Principal Component Analysis (PCA)~\cite{abdi2010principal} (RdimKD-P), Autoencoder (RdimKD-A), and Random Orthogonal Matrices~\cite{bingham2001random,johnson1984extensions} (RdimKD-R), explained in detail in~\cref{method}.

Our main highlights are summarized as follows: 
\begin{itemize}
\item Compared with previous methods that manipulate or align features in elaborate and fancy ways, this work reveals the benefit of dimensionality reduction in distillation from an essential level.
\item RdimKD works well on various deep learning tasks (image classification, object detection, semantic segmentation, language understanding, speech recognition) and neural architectures (CNNs, Transformer, Conformer), which makes it scalable to complex and diverse industrial applications.
\item Experiments show that RdimKD achieves performance comparable to or higher than state-of-the-art results on the above benchmarks.
\item The implementation of RdimKD, especially RdimKD-R is very simple yet effective, and has been landed in one of the most famous short-video companies, which means that it has been evaluated in the practice of super large-scale industry projects.
\end{itemize}

\section{Related works}
Knowledge distillation (KD) was first proposed by Hinton \etal~\cite{2015Distilling} in classification, where they utilize the logits from the teacher as the soft labels to transfer the ``dark knowledge'' to the student. Later, Fitnets~\cite{romero2014fitnets} started to distill knowledge from the intermediate layers to further boost the performance of students. Since then,  the mainstream KD methods can be roughly divided into logits distillation~\cite{2015Distilling, cho2019efficacy, furlanello2018born,mirzadeh2020improved,yang2019snapshot,zhang2018deep, zhao2022decoupled,stanton2021does,wang2018kdgan,phuong2019towards,beyer2022knowledge,liu2019knowledge,park2021prune} and intermediate layer distillation~\cite{romero2014fitnets,li2017mimicking,zagoruyko2016paying,park2019relational,yang2022masked,heo2019comprehensive,huang2017like,peng2019correlation,yim2017gift,tian2019contrastive,liu2019structured,ye2022generalized,yang2022mixskd,song2022spot,chen2021cross,tung2019similarity}. RdimKD falls into the latter category, and we will summarize related works in this category.


Distillation from intermediate layers can be considered as a regularization of models training. Besides classification, some works are designed for detection~\cite{chen2017learning,li2017mimicking,wang2019distilling,guo2021distilling,dai2021general,zhang2020improve,li2022knowledge2}, segmentation\cite{he2019knowledge,liu2019structured,yang2022cross,wang2020intra}, and other specific domains~\cite{pan2020spatio,wang2019progressive,xu2022mind,ding2022kd,wu2022tinyvit,chen2022dearkd,zhang2022wavelet}. In our paper, we want to construct a general distillation method for all these tasks. SAKD~\cite{song2022spot} proposed a strategy to adaptively determine the distillation layers in the teacher per sample in each training iteration during the distillation period. ReviewKD~\cite{chen2021distilling} built connection paths across different levels between teacher and student. MGD~\cite{yang2022masked} utilized a mask and some transformation modules to make the feature maps of student mimic that of teacher. CD~\cite{shu2021channel} normalized the feature map of each channel to obtain a distribution, then minimizes the Kullback–Leibler (KL) divergence between the distribution of teacher and student. TAT~\cite{lin2022knowledge} proposed a one-to-all method that allows each pixel of the teacher to teach all spatial locations of student given the similarity. Most of these methods focus on specific view of teacher's feature map by using some elaborated transformation, but failed to capture the generic information of KD. Rather than simply proposing another variant like before, we abstract a more generic and higher level perspective to reveal the nature of the problem.

\section{Methods}
\label{method}
We introduce the details of RdimKD in this section, including the overall design and the three projection matrices. 
The overall framework is shown in~\cref{fig:network}. Taking CNN as an example, let $F_t, F_s \in \mathbb{R}^{b\times h \times w \times c}$ be the feature maps to be distilled of the teacher and student, respectively. They can be viewed as two matrices, $F_t, F_s \in \mathbb{R}^{N \times c}$ representing a collection of $N$ points with $c$ dimensions (where $N=bhw$, and the same notation $F_t, F_s$ is used when not confused). These two matrices can be multiplied by a common fixed matrix $K \in \mathbb{R}^{c \times d}(d<c)$ to project these points into a subspace with $d$ dimensions.
In the subspace, the simple $\ell^2$ loss is used to minimize the difference between the two projected feature maps. 
The final objective function is as follows:
\begin{equation}
    \min_{w}\mathcal{F}(w) = \mathcal{L}(w)+ \frac{\alpha}{Nd}\|F_tK-F_sK\|^2
    \label{eq:objective}
\end{equation}
where $w$ is all learnable parameters of the student, $\mathcal{L}(w)$ the original loss function of the student, $\alpha$ the balance factor. RdimKD focuses on the concept of dimensionality reduction itself, regardless of the specific reduction methods.
To show this, we provide three methods to construct the projection matrix $K$, explained in the following subsections. Unlike learnable modules of some previous works, we freeze $K$ during the whole training process. Also, we will study the performance when it is changeable at each iteration in the ablation study part.

\subsection{Projection via PCA}
Principal component analysis (PCA)~\cite{abdi2010principal} is a popular technique for reducing the dimensionality of a dataset such that the variance of the dataset is preserved as much as possible. Specifically, 
we first center the values of each point in $F_t$ by subtracting the mean of each column from each of those values, resulting in matrix $\hat{F_t}$.  
The eigenvalue decomposition of its covariance matrix, $\frac{1}{N-1}\hat{F_t}^T\hat{F_t}$, is as follows:
\begin{equation}
    \frac{1}{N-1}\hat{F_t}^T\hat{F_t}=U\Sigma U^T
\end{equation}
where $\Sigma=diag\{\sigma_1, \sigma_2, ..., \sigma_c\}$ is a diagonal matrix, and each entry represents an eigenvalue. For the sake of description, we assume that they are already in descending order. $U=(u_1, u_2, ..., u_c)$ is the matrix whose columns $u_i, i=1,2,...,c$ are units and orthogonal to each other. Each of $u_i$ can be interpreted as a principal axis, and $\sigma_i$ is the corresponding variance along the $i$-th axis.

\begin{figure}[t]
  \includegraphics[width = 0.22\textwidth]{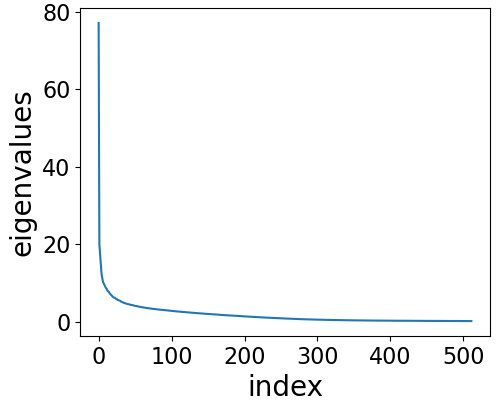}
  
  \vspace*{\dimexpr-\parskip-8\baselineskip}
  \parshape 9 
    .25\textwidth 0.25\textwidth 
    .25\textwidth 0.25\textwidth 
    .25\textwidth 0.25\textwidth
    .25\textwidth 0.25\textwidth
    .25\textwidth 0.25\textwidth
    .25\textwidth 0.25\textwidth
    .25\textwidth 0.25\textwidth
    .25\textwidth 0.25\textwidth
    0pt 0.5\textwidth 
  \makeatletter
  \refstepcounter\@captype
  \addcontentsline{\csname ext@\@captype\endcsname}{\@captype}
    {\protect\numberline{\csname the\@captype\endcsname}{ToC entry}}%
  \csname fnum@\@captype\endcsname: 
  \makeatother
  \small{The distribution of eigenvalues of the covariance matrix of the last feature maps in the fourth stage for ResNet-34 on ImageNet~\cite{deng2009imagenet}. We can see that the distribution is very anisotropic. Similar phenomena have also occurred in other scenarios.}
  \label{fig:eigen}
\end{figure}


As an example,~\cref{fig:eigen} shows the distribution of eigenvalues for ResNet-34 on ImageNet~\cite{deng2009imagenet}. The severe anisotropy of the distribution suggests that we may only need to project the samples to the first $d$ principal axes to represent the most important information. Hence, we can let $K=(u_1, u_2, ..., u_d)$ in~\cref{eq:objective}. The distillation method corresponding to the projection matrix obtained in this way is named RdimKD-P. We will show in the experiment section that projecting to the first $d$ principal axes does give better performance than projecting to the last $d$ principal axes.

\subsection{Projection via autoencoder}
PCA aims at projecting the dataset into a normal subspace while preserving the maximum amount of information. Another way to remove noise while retaining the primary information is to use an autoencoder. The matrix $K$ in~\cref{eq:objective} can be viewed as an encoder, and we design $K' \in \mathbb{R}^{d \times c}$ as a decoder, to minimize the objective function:
\begin{footnotesize}
\begin{equation}
    \min_{K, K'}\mathcal{J}(K, K') = \frac{1}{Nc}\|F_t - F_tKK'\|^2+\gamma(\|K\|^2+\|K'\|^2)
    \label{eq:objective_autoencoder}
\end{equation}
\end{footnotesize}
where $\gamma$ is a small positive number to balance the norm of $K$ and $K'$. Since the decoder $K'$ is used to restore the original information as much as possible, the encoded feature map $F_tK$ needs to retain general information of $F_t$.
When this is done, the solution of $K$ will act as the projection matrix in~\cref{eq:objective}. We name this method RdimKD-A.

\subsection{Projection via random orthogonal matrix}
Random projection~\cite{achlioptas2001database} is a technique used to reduce the dimensionality of datasets in Euclidean space. The core idea behind it is given in the Johnson-Lindenstrauss (JL) lemma~\cite{johnson1984extensions,matouvsek2008variants}:
\begin{lemma}
For any $0< \epsilon < 1$ and integer $N$, let $d$ be an integer with $d>4(\epsilon^{2}/2-\epsilon^3/3)^{-1}{\rm log} N$. Then, for any set $V$ of $N$ points in $\mathbb{R}^c$, there is a linear map f: $\mathbb{R}^c\rightarrow\mathbb{R}^d$ such that for all $u,v \in V$, the inequality holds:
\begin{equation}
    (1-\epsilon)\|u-v\|^2 \le \|f(u)-f(v)\|^2 \le  (1+\epsilon)\|u-v\|^2
\end{equation}
\label{lamma}
\end{lemma}
One proof takes $f$ to be a suitable multiple of orthogonal projector onto a random subspace, and it can be easily proved in~\cite{dasgupta2003elementary}. This lemma states that datasets in the space of high dimension can be linearly projected onto low-dimensional space with approximate preservation of distances between the samples. Random projection is simple and computationally efficient compared with other dimensionality reduction methods. We find that this idea can also be borrowed in the field of knowledge distillation, although the dimensions before and after projection do not strictly satisfy the requirements of the JL lemma. By this idea, we generate the random matrix $K$ in~\cref{eq:objective} in the following two steps: 1. generating a random matrix in $\mathbb{R}^{c \times d}$ with elements chosen from Gaussian distribution; 2. orthonormalizing all the columns of the matrix by Gram–Schmidt process. These two steps are very simple to implement in PyTorch~\cite{paszke2019pytorch} with just one line of code:
{\small \verb|torch.nn.init.orthogonal_|}

Matrix obtained in this way has spherical symmetry, which we guess may be a good property. It is possible that in some extreme cases, the projection matrix will project the samples to a subspace that approximates the span of the last $d$ principal axes of the PCA, which will result in bad performance shown in~\cref{tab:ablation_projection_method_resnet34} pac\_last. Nevertheless, at least we did not observe this extreme phenomenon in our experiments. We name this method RdimKD-R.

\section{Experiments}
\label{Experiments}
To show the generality and effectiveness of RdimKD, we conduct experiments on various deep learning tasks (image classification, object detection, semantic segmentation, language understanding, and speech recognition) and neural architectures (CNN~\cite{he2016deep}, Transformer~\cite{vaswani2017attention}, and Conformer~\cite{gulati2020conformer}), and compare them to works in recent years. We set $r=c/d$, which represents the reduction rate of the subspace dimension. For simplicity, we use the same $r$ for all feature maps to be distilled given an experiment. For RdimKD-A, we choose to train~\cref{eq:objective_autoencoder} by gradient descent before distillation training, although it may have a closed-form solution. For RdimKD-P, we randomly selected hundreds of training samples to conduct PCA. In the following, for a brief description, ``A to B" means a distillation experiment with A as the teacher and B as the student. Due to the page limit, we only explain the primary settings for some experiments, and
other details and language understanding are attached in the supplementary materials.

\subsection{Image classification}
The classification experiments are done on ImageNet ILSVRC-12 dataset~\cite{deng2009imagenet}, which contains 1000 object categories with 1.2 million images for training and 50k for testing. We conduct experiments on ``ResNet-34 to ResNet-18" and ``ResNet-50 to MobileNet~\cite{howard2017mobilenets}''. Top-1 accuracy is reported. RdimKD can be combined with solf label based KD~\cite{2015Distilling}, that is, by adding the additional loss:
\begin{equation}
    \mathcal{L}_{KL}=-\beta \sum_iq_i \log p_i
\end{equation}
where $q_i$ is the probability distribution of the teacher's output, $p_i$ is that of the student, and $\beta$ is the balance coefficient. 

ResNet-34 and ResNet-18 contain four stages, and the difference is that the number of blocks in each stage is different. We distilled the last feature map of the third and fourth stages, where the number of channels is 128 and 512, respectively. For MobileNet, we distill the last feature maps of the third and fourth stages of Resnet-50 to the outputs of the 11th and 14th convolutions of MobielNet. To achieve this, we manually changed the number of channels in these two layers of ResNet-50 to 512 and 1024, respectively, and the necessary 1x1 convolution is added at the skip layer.
Data preprocessing and augmentation are the same as that of PyTorch official example\footnote{https://github.com/pytorch/examples/blob/master/imagenet/main.py}. We use a cosine learning rate scheduler with an initial value of 0.1 and train it for 105 epochs. In ``ResNet-34 to ResNet-18'', $r=4$ and $\alpha=1$ for all the RdimKD, and $\beta=2$ for RdimKD.
 Results are shown in~\cref{tab:res34to18}. We can see that our RdimKD can boost performance by a clear margin, and the simplest RdimKD-R produces about the same performance as RdimKD-A/P.

\begin{table}
\begin{center}
\footnotesize
  \setlength\tabcolsep{3pt}
  \renewcommand{\arraystretch}{1.1}

  \begin{tabular}{lll|lll}
\toprule
method    & MbNet & Res18 & method & MbNet & Res18 \\
\hline
teacher   & 76.77     & 74.55     & DIST~[NIPS 2022]~\cite{huang2022knowledge}   & \textbf{73.24}     & 72.07     \\
student   & 70.93     & 70.96     & WSLD~[ICLR 2021]~\cite{zhou2021rethinking}   & 71.52     & 72.04     \\
RdimKD-R  & 72.56     & 71.89     & SRRL~[ICLR 2021]~\cite{yang2021knowledge}   & 72.49     & 71.73     \\
RdimKD-A  & 72.65     & 71.94     & KR~[CVPR 2021]~\cite{chen2021distilling}     & 72.56     & 71.61     \\
RdimKD-P  & 72.77     & 72.01     & DKD~[CVPR2022]~\cite{zhao2022decoupled}    & 72.05     & 71.7      \\
RdimKD-R* & 73.13     & 72.53     & MGD~[ECCV 2022]~\cite{yang2022masked}    & 72.59     & 71.8      \\
RdimKD-A* & 73.15     & \textbf{72.58}     & TAT~[CVPR 2022]~\cite{lin2022knowledge}    & None      & 72.41     \\
RdimKD-P* & 73.23     & 72.49     & KCD~[ECCV 2022]~\cite{li2022knowledge}    & 71.25     & 72.13     \\
    \bottomrule
  \end{tabular}
  \end{center}
  \caption{Top-1 results on ImageNet. For MbNet column, ResNet-50 is teacher and MobileNet is student; while for Res18 column, ResNet-34 is teacher and ResNet-18 is student. * means combined with $\mathcal{L}_{KL}$, and None means not reported in origin paper. Note that TAT, KCD, DIST, WSLD also contain $\mathcal{L}_{KL}$. We can see that RdimKD can boost the performance of student by a clear margin, and that the simplest RdimKD-R can get comparable performance as RdimKD-A/P. All of our results are the average on 3 trials.}
  \label{tab:res34to18}
\end{table}

\subsection{Object detection}
The detection experiments are conducted on the COCO2017 dataset~\cite{lin2014microsoft}, which contains 80 object categories with 115k training images and 5k validation images. All the teachers are trained for 36 epochs and students for 24 epochs. Other training details are the same as the standard protocols in the widely used Detectron2 library~\cite{wu2019detectron2}. Inheriting strategy~\cite{kang2021instance} is used for distillation. 
The networks are RetinaNet~\cite{lin2017focal} and Faster-RCNN~\cite{ren2015faster} with different backbones. Mean average precision(AP) is reported.

\textbf{RetinaNet: } RetinaNet uses a Feature Pyramid Network (FPN)~\cite{lin2017feature} to generate a multi-scale feature pyramid with levels $P_3$ to $P_7$, all of which contain 256 channels. The only difference between the various benchmark models is the backbone. So, naturally, knowledge can be distilled from the five levels of the pyramid. We use two teacher networks, RetinaNet-ResNet-101 and RetinaNet-ResNeXt-101~\cite{xie2017aggregated}, separately, to teach the student network, RetinaNet-ResNet-50. Results are shown in~\cref{tab:retinaNet}. In these experiments, $r=4$. For RdimKD-R and RdimKD-A, $\alpha=1$, while for RdimKD-P, $\alpha=0.5$.

\textbf{Faster-RCNN: } Besides the one-stage detector RetinaNet, we also evaluate our RdimKD in two-stage detector, Faster-RCNN. Similar to previous works~\cite{zhao2022decoupled, yang2022focal}, we also use FPN to capture multi-scale features. Same as DKD~\cite{zhao2022decoupled}, $P_2$ to $P_5$ are input features for the ROI heads, and the number of channels for each level is 256. Similar to RetinaNet, the only difference between various benchmarks is the backbone, while the structure of the ROI head is the same. So, naturally, we distill at the FPN layers. In these experiments, we conduct ``Faster-RCNN-FPN-ResNet-101 to Faster-RCNN-FPN-ResNet-18'', ``Faster-RCNN-FPN-ResNet-101 to Faster-RCNN-FPN-ResNet-50'' and ``Faster-RCNN-FPN-ResNet-101 to Faster-RCNN-FPN-MobileNet\_V2~\cite{sandler2018mobilenetv2}''. $r=4$ for ResNet-50 and MobileNet\_V2, $r=2$ for ResNet-18. Details are in Supplementary Materials. Results are shown in~\cref{tab:faster_rcnn}. 

\begin{table}
\begin{center}
   \small
  \setlength\tabcolsep{5pt}
  \begin{tabular}{l|llll}
    \toprule
method            & AP    & $\text{AP}_\text{S}$   & $\text{AP}_\text{M}$   & $\text{AP}_\text{L}$   \\
\hline
S:RetinaNet-R50  & 38.28 & 22.36 & 42.34 & 49.42 \\
\hline
\hline
T:RetinaNet-R101 & 40.56 & 24.46 & 44.57 & 52.73 \\
\hline
FGD~[CVPR 2022]~\cite{yang2022focal}               & 39.7  & 22.0  & 43.7  & 53.6  \\
KR~[CVPR 2021]~\cite{chen2021distilling}          & 38.48 & 22.67 & 42.72 & 58.22 \\
LD~[CVPR 2022]~\cite{zheng2022localization}                & 39.0  & 23.1  & 43.2  & 51.1  \\
FRS~[NIPS 2021]~\cite{zhixing2021distilling}               & 39.7  & 21.8  & 43.5  & 52.4  \\
LGD~[AAAI 2022]~\cite{zhang2022lgd} & 40.35 & 24.08 & 44.15 & 52.53\\ 
GID~[CVPR 2021]~\cite{dai2021general} & 39.1 & 22.8 & 43.1 & 52.3 \\
KDRP~[AAAI 2022]~\cite{li2022knowledge2} & 39.6 & 21.4 & 44.0 & 52.5 \\
RdimKD-R          & 40.67 & 24.57 & 44.62 & 52.92 \\
RdimKD-A          & 40.67 & 24.45 & 44.70 & 53.16 \\
RdimKD-P          & \textbf{40.68} & 24.17 & 44.90 & 52.72 \\
\hline
\hline
T:RetinaNet-X101 & 41.10 & 23.95 & 44.78 & 53.27 \\
\hline
FGD~[CVPR 2022]~\cite{yang2022focal}               & 40.7  & 22.9  & 45.0  & 54.7  \\
MGD~[ECCV 2022]~\cite{yang2022masked}               & 41.0  & 23.4  & 45.3  & 55.7  \\
FRS~[NIPS 2021]~\cite{zhixing2021distilling}               & 40.1  & 21.9  & 43.7  & 54.3  \\
DIST~[NIPS 2022]~\cite{huang2022knowledge}              & 40.1  & 23.2  & 44.0  & 53.6  \\
CD~[ICCV 2021]~\cite{shu2021channel}                & 40.8  & 22.7  & 44.5  & 55.3  \\
FB~[ICLR 2021]~\cite{zhang2020improve}                & 39.6  & 22.7  & 43.3  & 52.5  \\
LGD~[AAAI 2022]~\cite{zhang2022lgd} & 40.35 & 24.08 & 44.15 & 52.53\\ 
RdimKD-R          & 40.97 & 24.20 & 45.18 & 53.99 \\
RdimKD-A          & 40.95 & 23.72 & 45.11 & 53.86 \\
RdimKD-P          & \textbf{41.05} & 24.47 & 45.39 & 54.02 \\
    \bottomrule
  \end{tabular}
  \end{center}
  \caption{Performance of ``RetinaNet-R101 to RetinaNet-R50''(the top half of the table) and ``RetinaNet-X101 to RetinaNet-R50''(the bottom half of the table) on COCO2017 validation set. Where `R101',`R50' and `X101' mean ResNet-101, ResNet-50 and ResNeXt-101, respectively. `T' and `S' mean teacher and student, respectively. LGD~\cite{zhang2022lgd} is a self-distillation method and does not contain teachers. We can see that RdimKD consistently equals or outperforms other methods. The simplest RdimKD-R produces about the same performance as RdimKD-A/P. All of our results are the average on 3 trials.}
  \label{tab:retinaNet}
\end{table}

\begin{table}
\begin{center}
\small
  \setlength\tabcolsep{5pt}
  \begin{tabular}{l|llll}
    \toprule
method   & AP    & $\text{AP}_\text{S}$   & $\text{AP}_\text{M}$   & $\text{AP}_\text{L}$   \\
\hline
T: ResNet-101  & 42.17 & 25.50 & 45.55 & 54.93 \\
S: ResNet-18   & 34.96 & 19.78 & 37.39 & 45.44 \\
\hline
DKD~[CVPR 2022]~\cite{zhao2022decoupled}      & 37.01 & None     & None     & None     \\
SCKD~[ICCV 2021]~\cite{zhu2021student}     & 37.5  & 20.9  & 42.6  & 50.8  \\
KR~[CVPR 2021]~\cite{chen2021distilling}       & 36.75 & 19.42 & 39.51 & 49.58 \\
RdimKD-R & 38.25 & 21.12 & 41.25 & 51.34 \\
RdimKD-A & 38.29 & 21.04 & 41.03 & 51.45 \\
RdimKD-P & \textbf{38.31} & 20.78 & 41.12 & 51.82 \\
\hline
\hline
T: ResNet-101  & 42.17 & 25.50 & 45.55 & 54.93 \\
S: ResNet-50   & 39.66 & 24.03 & 42.76 & 51.74 \\
\hline
DKD~[CVPR2022]~\cite{zhao2022decoupled}      & 40.65 & None     & None     & None     \\
FGD~[CVPR 2022]~\cite{yang2022focal}      & 40.5  & 22.6  & 44.7  & 53.2  \\
ICD~[NIPS 2021]~\cite{kang2021instance}      & 40.9  & 24.5  & 44.2  & 53.5  \\
KR~[CVPR 2021]~\cite{chen2021distilling}       & 40.36 & 23.60 & 43.81 & 52.87 \\
FRS~[NIPS 2021]~\cite{zhixing2021distilling}      & 39.5  & 22.3  & 43.6  & 51.7  \\
LGD~[AAAI 2022]~\cite{zhang2022lgd}      & 40.47 & 23.96 & 43.94 & 52.19 \\
RdimKD-R & \textbf{41.76} & 25.22 & 45.28 & 54.42 \\
RdimKD-A & 41.72 & 25.12 & 45.11 & 54.50 \\
RdimKD-P & 41.74 & 24.97 & 45.18 & 54.60 \\
\hline
\hline
T: ResNet-50   & 41.84 & 25.15 & 45.27 & 54.48 \\
S: MobileNet\_V2   & 34.51 & 19.98 & 36.38 & 44.94 \\
\hline
DKD~[CVPR2022]~\cite{zhao2022decoupled}      & 34.35 & None     & None     & None     \\
KR~[CVPR 2021]~\cite{chen2021distilling}       & 33.71 & 16.77 & 35.81 & 46.47 \\
RdimKD-R & \textbf{36.26} & 20.24 & 38.29 & 49.21 \\
RdimKD-A & 36.06 & 19.82 & 37.95 & 48.86 \\
RdimKD-P & 36.00 & 19.50 & 38.00 & 49.20 \\
    \bottomrule
  \end{tabular}
  \end{center}
  \caption{Results on COCO2017 on Faster-RCNN-FPN with different backbones. `T' and `S' mean teacher and student, respectively. In the table, the top part is for ``Faster-RCNN-FPN-ResNet-101 to Faster-RCNN-FPN-ResNet-18'', the middle part is for ``Faster-RCNN-FPN-ResNet-101 to Faster-RCNN-FPN-ResNet-50'', while the bottom part is for ``Faster-RCNN-FPN-ResNet-101 to Faster-RCNN-FPN-MobileNet\_V2''. `None' means not reported in the original paper. For MobileNet\_V2, the baseline for DKD~\cite{zhao2022decoupled} and KR~\cite{chen2021distilling} is relatively weak, such that our baseline performance, 34.51, is higher than that of DKD and KR after their distillation. We choose the best of all the ResNet-50, under the guidance of ResNet-101 with RdimKD-R, as the teacher for MobileNet\_V2. All the results are average on 3 trials.}
  \label{tab:faster_rcnn}
\end{table}

\begin{table}
\begin{center}

\small

  \centering
  \setlength\tabcolsep{5pt}
  \begin{tabular}{l|cc}
    \toprule
method   & ResNet18 & MobileNet\_V2 \\
\hline
teacher  & 79.26    & 79.26         \\
student  & 73.59    & 73.70         \\
\hline
TAT~[CVPR 2022]~\cite{lin2022knowledge}      & 75.76    & 73.85         \\
CIRKD~[CVPR 2022]~\cite{yang2022cross}    & 74.50    & None          \\
FAKD~\cite{yuan2022fakd}     & None     & 67.62         \\
IFVD~[ECCV 2020]~\cite{wang2020intra}     & 74.05    & None          \\
ICKD~[ICCV 2021]~\cite{liu2021exploring}     & 75.01    & 72.79         \\
RdimKD-R & 75.63    & 75.28         \\
RdimKD-A & 75.55    & \textbf{75.67}         \\
RdimKD-P & \textbf{75.94}    & 75.19     \\

    \bottomrule
  \end{tabular}
  \end{center}
  \caption{Results on Semantic segmentation on Pascal VOC. We use DeepLabv3+-ResNet-101 as teacher, and DeepLabv3+-ResNet-18 and DeepLabv3+-MobileNet\_V2 as sudents. `None’ means not reported in the original paper. Also, for MobileNet\_V2, the baseline for FAKD is relatively weak. All of our results are the average on 6 trials.}
  \label{tab:Semantic segmentation}
\end{table}

\subsection{Semantic segmentation}
The segmentation experiments are done on the Pascal VOC~\cite{everingham2010pascal}, which contains 20 foreground classes and 1 background class. With the additional coarse annotated training images from~\cite{hariharan2011semantic}, there are a total of 10582 images for training. The validation set contains 1499 images, on which we report the mean Intersection over Union (mIoU) to show segmentation performance.

\textbf{DeepLabv3+: } DeepLabv3+~\cite{chen2018encoder} is a popular network for segmentation tasks. Besides the Atrous Spatial Pyramid Pooling (ASPP) module and encode-decoder structure, it extends DeepLabv3~\cite{chen2017rethinking} by adding a decoder module to capture rich semantic information to refine the object boundaries. Knowledge is distilled at the low-level feature coming from the backbone (in front of the resize and ReLU layer) and the output of the ASPP module (in front of the final dropout and ReLU layer), where the number of channels is 48 and 256, respectively. We use the settings from the public code~\footnote{https://github.com/VainF/DeepLabV3Plus-Pytorch} unless otherwise stated. For all of our experiments, the output stride (OS) is 16 for training and validation. We conduct ``DeepLabv3+-ResNet-101 to DeepLabv3+-ResNet-18'' and ``DeepLabv3+-ResNet-101 to DeepLabv3+-MobileNet\_V2''. The values for $\alpha$ and $r$ are in supplementary materials. We find that the variance of the results for each trial is large, so each of our results is the average of the six trials. The results are shown in ~\cref{tab:Semantic segmentation}.

\subsection{Speech recognition}
To show the powerful generalization and effectiveness of RdimKD, we apply RdimKD to a more challenging task, speech recognition. The input to this task is a speech waveform, and the output is the corresponding text. RNN-Transducer (RNN-T)~\cite{graves2012sequence} is a well-known end-to-end architecture for streaming speech recognition~\cite{he2019streaming}, which contains an encoder, a predictor, and a joiner. For the encoder, we use 12 layers of Conformer~\cite{gulati2020conformer} for the teacher and 6 layers for the student. For the predictor, three layers of bidirectional transformers~\cite{vaswani2017attention} (Bitransformers) is used for both teacher and student. We use the Librispeech dataset~\cite{panayotov2015librispeech}, which contains about 1000 hours of speech sampled at 16 kHz. We use the 960 hours of corpus for training and the development and test set for evaluation. Beam search is used as the decode mode, and the Word Error Rate (WER) is used as the metric. The implementation details are the same as the corresponding part of Wenet Library~\cite{yao2021wenet,zhang2022wenet}. We trained 50 epochs for each experiment.
The dimension of attention for the encoder is 256, and we distill knowledge from the 8th and 12th layers of the teacher to the 4th and 6th layers of the student. $\{\alpha=2, r =4\}$ is chosen for all results in this experiment. Inheriting strategy~\cite{kang2021instance} is used. The results are shown in~\cref{tab:librispeech}. It can be seen that our RdimKD has strong generalization and effectiveness besides computer vision.

\begin{table}
\begin{center}
  \setlength\tabcolsep{5pt}
\begin{tabular}{l|ccc|ccc}
\toprule
\multicolumn{1}{c|}{\multirow{2}{*}{}} & \multicolumn{3}{c|}{clean(\%)} & \multicolumn{3}{c}{other(\%)} \\
\cline{2-7}
\multicolumn{1}{c|}{}                  & dev     & test   & mean   & dev     & test   & mean   \\
\hline
Teacher                               & 3.30    & 3.48   & 3.39   & 8.86    & 8.90   & 8.88   \\
Student                               & 3.71    & 3.99   & 3.85   & 10.22   & 10.21  & 10.22  \\
\hline
RdimKD-R                              & 3.51    & 3.80   & \textbf{3.66}   & 9.40    & 9.50   & 9.45   \\
RdimKD-A                              & 3.56    & 3.78   & 3.67   & 9.44    & 9.44   & \textbf{9.44}   \\
RdimKD-P                              & 3.56    & 3.77   & 3.67   & 9.48    & 9.43   & 9.46   \\
no\_proj                              & 3.67    & 3.98   & 3.83   & 10.10    & 10.02   & 10.06   \\
\bottomrule
\end{tabular}
\end{center}
  \caption{Results on speech recognition. We use 12 layers of encoder of RNN-T as teacher and 6 as student, and use Librispeech as experiment dataset. Word Error Rate (WER) is used as metric (smaller is better). no\_proj is explained in~\cref{noproj}. The mean column is the arithmetic mean of dev and test columns. It can be seen that our RdimKD has strong generalization and effectiveness besides computer vision. Only one trial is run for each result in this table due to the high cost of the experiment. 
  }
  \label{tab:librispeech}
\end{table}

\subsection{Ablation studies}
The key to RdimKD is the projection. In this subsection, we focus on the projection methods (including no projection), subspace reduction rate $r$, and the coefficient of distillation loss $\alpha$. For generality, we will explore these ablation studies in different experiments. 
\label{noproj}

\textbf{Projection method: } As shown in 
~\cref{tab:ablation_projection_method_resnet34}, we conduct this ablation study via ``ResNet-50 to MobileNet” on ImageNet, ``Faster-RCNN-FPN-ResNet-101 to Faster-RCNN-FPN-ResNet-18” on COCO2017, and ``DeepLabv3+-ResNet-101 to DeepLabv3+-MobileNet\_V2'' on Pascal VOC. We have the following findings: 1. In RdimKD-P, feature maps are projected to PCA's first $d$ principal axes. A natural question is what will happen if feature maps are projected to the last $d$ principal axes (named by pca\_last in~\cref{tab:ablation_projection_method_resnet34}). Comparing the results between RdimKD-P and pca\_last, we find that the first $d$ principal components do contain more valuable information. Theoretically, in an extreme case, the projection matrix of RdimKD-R approximates that of pca\_last. Nevertheless, RdimKD-R consistently performed well in our experiments. 2. If we remove the projection operation and apply the $\ell^2$ loss directly onto the original feature maps (named by no\_proj in~\cref{tab:ablation_projection_method_resnet34}), the performance can be improved compared to the baseline. However, the improvement is smaller than that with projection (also shown in~\cref{tab:librispeech}) \textbf{consistently}. We suspect that this is caused by the low capacity of students, which makes it impossible and unnecessary to learn every detail from the teacher network accurately. 3. When the random projection matrix is not necessarily orthogonal (each element is chosen from Gaussian distribution $\mathcal N (0, \frac{1}{c})$, named by no\_orth), the performance is slightly worse than RdimKD-R. 4. If the random projection matrix is generated in every iteration rather than kept fixed from the beginning(named by randE), the performance is also slightly worse than RdimKD-R. It should be noted that the above comparison is not necessarily fair because each method may correspond to a unique optimal $\alpha$ value, and it is difficult to find the optimal $\alpha$ for each method due to the experimental cost. Nonetheless, RdimKD performs better than no\_proj and RdimKD-P performs better than pca\_last for a certain range of $\alpha$.

\begin{figure}
\centering
\subfigure[ImageNet]{
\includegraphics[width = 0.21\textwidth]{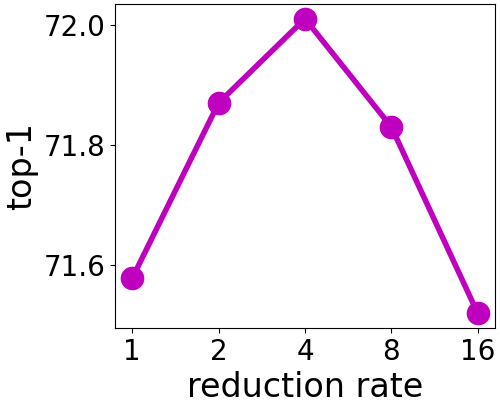}
\label{fig:r_res34t18_r}
}
\subfigure[COCO]{
\includegraphics[width = 0.21\textwidth]{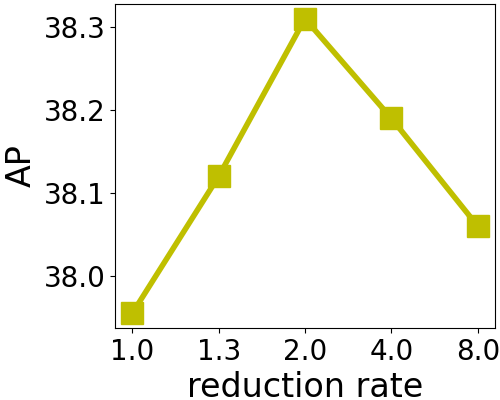}
\label{fig:r_faster}
}
\caption{Ablation study for reduction rate $r$. ~\cref{fig:r_res34t18_r} is RdimKD-P for ``ResNet-34 to ResNet-18'' on ImageNet classification, while ~\cref{fig:r_faster} is RdimKD-R for ``Faster-RCNN-FPN-ResNet-101 to Faster-RCNN-FPN-ResNet-18'' on COCO object detection. The X-axis is the reduction rate $r$, while the Y-axis is the performance (top-1 for ImageNet, AP for COCO). We can see that the projection of the feature maps onto a subspace of appropriate dimensions does boost the performance. All the results are average on 3 trials.}
\label{fig:reduction_rate}
\end{figure}

\textbf{Reduction rate $r$: }
The reduction rate is also an important variable, which determines the dimensional scaling of the subspace onto which the data of the original space is projected. 
It is easy to prove that $r=1$ is mathematically equivalent to no\_proj, and they are close in experimental performance.
As shown in~\cref{fig:reduction_rate}, we can see that the projection of feature maps onto a subspace of appropriate dimensions does boost performance further.

\begin{table}
\begin{center}
  \vspace{-0.1in}
  \setlength\tabcolsep{5pt}
  \begin{tabular}{lcc|lcc}
    \toprule
    method   & top-1        & top-5      & method    & top-1        & top-5      \\
    \hline
baseline & 70.93        & 89.59      & pca\_last & 71.40        & 89.96      \\
RdimKD-R & 72.56        & 90.94      & no\_proj  & 72.24        & 90.79      \\
RdimKD-A & 72.65        & 91.00      & no\_orth  & 72.40        & 90.87      \\
RdimKD-P & 72.77        & 91.02      & randE   & 71.63        & 90.47       \\
\hline
\hline
method   & AP  & mIOU  & method    & AP & mIOU   \\
\hline
baseline & 34.96 & 73.70 & pca\_last & 35.89 & 74.40 \\
RdimKD-R & 38.31 & 75.28 & no\_proj  & 37.96 & 74.93 \\
RdimKD-A & 38.29 & 75.67 & no\_orth  & 37.75 & 75.04 \\
RdimKD-P & 38.31 & 75.19 & randE   & 37.77 & 74.91 \\
    \bottomrule
  \end{tabular}
  \end{center}
  \caption{Ablation study for different projection methods. The top part is ``ResNet-50 to MobileNet'' on ImageNet, while the bottom is ``Faster-RCNN-FPN-ResNet-101 to Faster-RCNN-FPN-ResNet-18'' on COCO2017 (the AP column) and ``DeepLabv3+-ResNet-101 to DeepLabv3+-MobileNet\_V2'' on Pascal VOC (the mIOU column). pca\_last means feature maps are projected to the last $d$ principal axes of PCA. no\_proj means the $K$ in~\cref{eq:objective} is an identity matrix. no\_orth means that elements in the projection matrix $K$ in~\cref{eq:objective} are randomly chosen from Gaussian distribution $\mathcal N (0, \frac{1}{c})$, and the matrix itself is not necessarily orthogonal. randE means that the matrix $K$ is a \textbf{rand}om orthogonal matrix generated in \textbf{E}ach iteration, not fixed. Results for ImageNet and COCO2017 are average on 3 trials, and that for VOC are average on 6 trials.}
  \label{tab:ablation_projection_method_resnet34}
\end{table}

\begin{figure}
\centering
\subfigure[Pascal VOC]{
\includegraphics[width = 0.22\textwidth]{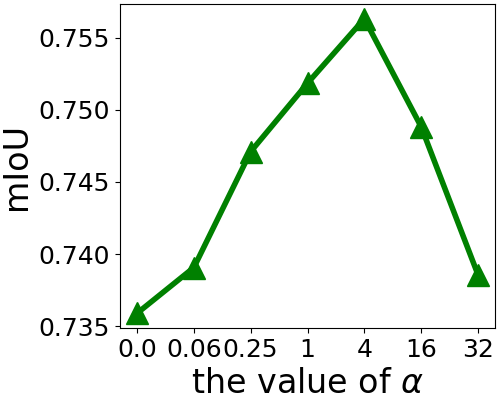}
\label{fig:alpha_res34t18}
}
\subfigure[ImageNet]{
\includegraphics[width = 0.22\textwidth]{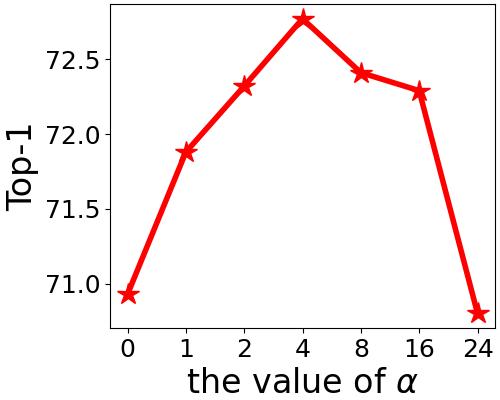}
\label{fig:alpha_alphares50tMB}
}
\caption{Ablation study for $\alpha$. The left is RdimKD-R for ``DeepLabv3+-ResNet-101 to DeepLabv3+-ResNet-18'' on the Pascal VOC segmentation task; while the right is RdimKD-P for ``ResNet-50 to MobileNet'' on ImageNet classification task. It can be shown that with the increase of $\alpha$ value, the student performance rises first and then decreases, which is in line with our expectations and verifies our method's effectiveness. The result for Pascal VOC is the average of 6 trials, and that for ImageNet is the average of 3 trials.}
\label{fig:alpha}
\end{figure}

\textbf{Coefficient $\alpha$: }
The value of $\alpha$ balances the weights between the original loss and the distillation loss. In~\cref{fig:alpha}, we implement ``DeepLabv3+-ResNet-101 to DeepLabv3+-ResNet-18'' on Pascal VOC segmentation task and ``ResNet-50 to MobileNet'' on ImageNet classification task. It can be shown that with the increase of $\alpha$ value, the performance of the student rises first and then decreases, which is in line with our expectations and verifies the effectiveness of our method.

\textbf{Distillation position: }
For non-sequential structures such as RetinaNet, we distill the feature maps of the output of FPN; for the sequential structure that stacks some blocks, layer indices to be distilled between teacher and student are proportional. For example, if teacher is 2 times deeper than the student, then the $i$-th layer of student is taught by the corresponding 2$i$-th layer of the teacher.
However, we find that, in general, only distilling some upper layers produces better results, as shown in~\cref{tab:distillation_position} as an example. 

\begin{table}
\begin{center}
  \setlength\tabcolsep{5pt}
\begin{tabular}{l|ccc|ccc}
\toprule
\multicolumn{1}{c|}{\multirow{2}{*}{mask}} & \multicolumn{3}{c|}{clean(\%)} & \multicolumn{3}{c}{other(\%)} \\
\cline{2-7}
\multicolumn{1}{c|}{}                  & dev     & test   & mean   & dev     & test   & mean   \\
\hline
111111                              & 3.58    & 3.79   & 3.69   & 9.65    & 9.78   & 9.72 \\
001111                              & 3.51    & 3.80   & 3.66   & 9.40    & 9.50   & 9.45  \\
000011                              & 3.54    & 3.78   & 3.66   & 9.47    & 9.53   & 9.50  \\
\bottomrule
\end{tabular}
\end{center}
  \caption{An example of distillation position for sequential structure network. 
  When 12-layer RNN-T teaching 6-layer, 
  the $i$-th layer of student is taught by the corresponding 2$i$-th layer of the teacher
  But not distillation the lower layers is better. Mask 001111 means not distilling the first and second layer of the student and 111111 means distilling all the 6 layers. The conclusion is consistent with~\cite{yang2022masked}.}
  \label{tab:distillation_position}
\end{table}

\subsection{Discussion}
RdimKD-R and RdimKD-P project the feature maps onto a subspace (denoted as $\mathcal{S}$), and an interesting point is the learning of student in the orthogonal complement (denoted as $\mathcal{S}^{\perp}$). Note that $\mathcal{S}$ and $\mathcal{S}^{\perp}$ are determined by the teacher. 

\begin{figure}
\centering
\subfigure[RdimKD-P]{
\includegraphics[width = 0.22\textwidth]{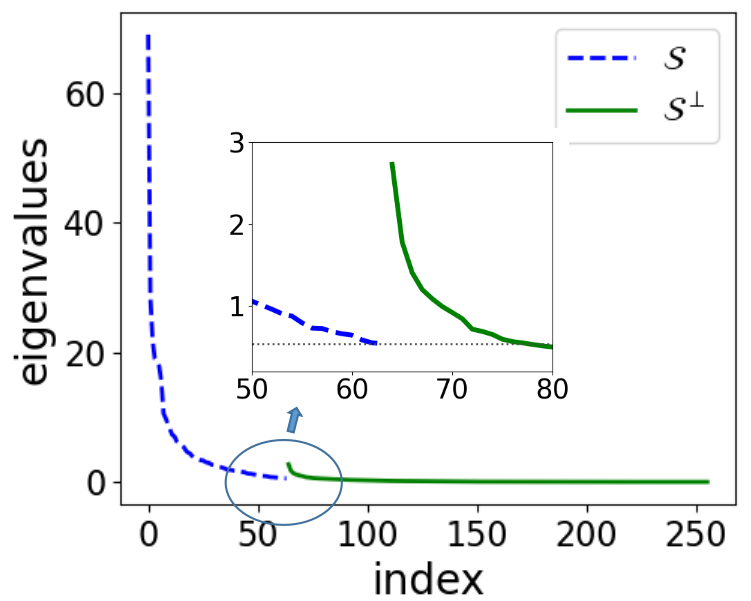}
\label{fig:discussion_pca}
}
\subfigure[RdimKD-R]{
\includegraphics[width = 0.22\textwidth]{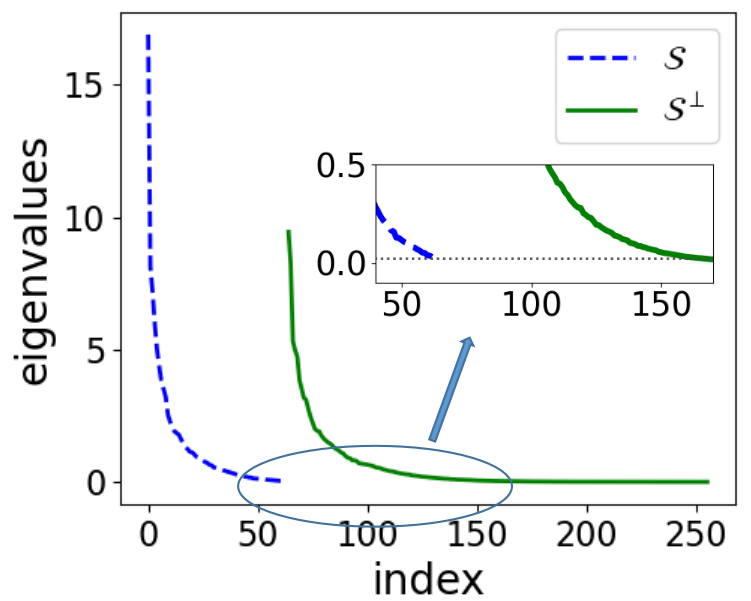}
\label{fig:discussion_random}
}
\caption{The distribution of eigenvalues of the covariance matrices of student's feature maps in $\mathcal{S}$ and $\mathcal{S}^{\perp}$. The experiment is done in ``Faster-RCNN-FPN-ResNet-101 to Faster-RCNN-FPN-ResNet-18'', and the feature map is the $P_5$ level. The left is by RdimKD-P and the right is by RdimKD-R. The eigenvalues with large index are very close to zero.}
\label{fig:discussion}
\end{figure}

For illustration, we name the subspace spanned by the first $d$ principal axes as the principal subspace. By this definition, for RdimKD-P, $\mathcal{S}$ is also the principal subspace of the teacher. A natural question is whether it is also the student's principal subspace after it is trained by RdimKD-P. To explore it, we project the feature map of the well-trained student into $\mathcal{S}$ and $\mathcal{S}^{\perp}$, respectively, and then do PCA in these two subspaces. The distribution of these eigenvalues is plotted in~\cref{fig:discussion_pca}. We find that the eigenvalues in $\mathcal{S}$ are almost all larger than those in $\mathcal{S}^{\perp}$, which shows that $\mathcal{S}$ is almost the principal subspace of the student, too. Interestingly, for RdimKD-R in~\cref{fig:discussion_random}, $\mathcal{S}$ is randomly chosen instead of by PCA, but the eigenvalues in $\mathcal{S}$ are also generally larger than those in $\mathcal{S}^{\perp}$. Comparing the ordinates of~\cref{fig:discussion_pca} and~\cref{fig:discussion_random}, the variance of the feature maps trained by RdimKD-R is smaller than that trained by RdimKD-P.

\begin{figure}
\centering
\subfigure[RdimKD-P]{
\includegraphics[width = 0.22\textwidth]{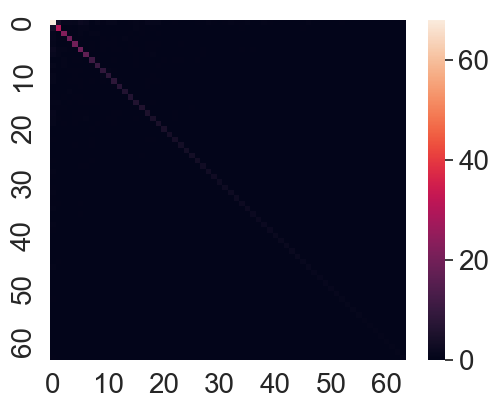}
\label{fig:rotate_pca}
}
\subfigure[RdimKD-R]{
\includegraphics[width = 0.22\textwidth]{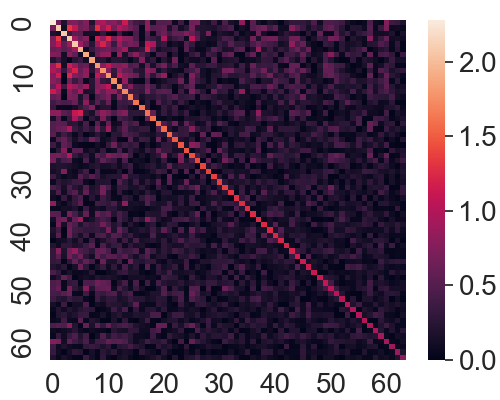}
\label{fig:rotate_random}
}
\caption{The heatmap of the absolute value of the covariance matrix. The experiment is done in ``Faster-RCNN-FPN-ResNet-101 to Faster-RCNN-FPN-ResNet-18'', and the feature map is the $P_5$ level. The left is by RdimKD-P and the right is by RdimKD-R. }
\label{fig:rotate}
\end{figure}

We also consider the rotation of the principal subspace of student with respect to that of the teacher. To do so, we first project the feature maps of the student onto $\mathcal{S}$ then plot the heat map of the covariance matrix in~\cref{fig:rotate_pca} (trained by RdimKD-P) and in~\cref{fig:rotate_random} (trained by RdimKD-R). 
From~\cref{fig:rotate_pca}, the value of the diagonal elements is much greater than the value of the off-diagonal elements, which leads to the conclusion that the angle between the two principal set of axes is small. For comparison, the heatmap via RdimKD-R is also plotted in~\cref{fig:rotate_random}.

\section{Conclusion}
In this paper, we proposed RdimKD with three projection methods for knowledge distillation. Compared with other methods, the advantage of RdimKD is three folds: simple to implement (especially RdimKD-R) and very favored for industrial applications; achieves performance comparable to or higher than state-of-the-art methods; general to various learning tasks and neural architectures. We believe our simple findings will bring more enlightenment and inspirations to knowledge distillation.
Moreover, this approach has been widely used in our company's industrial applications.
However, the theoretical explanation for why this so simple method works still needs further study.

\clearpage
{\small
\bibliographystyle{ieee_fullname}
\bibliography{egbib}
}

\end{document}


\title{Supplementary Materials: RdimKD: Generic Distillation Paradigm by Dimensionality Reduction}


\author{First Author\\
Institution1\\
Institution1 address\\
{\tt\small firstauthor@i1.org}
\and
Second Author\\
Institution2\\
First line of institution2 address\\
{\tt\small secondauthor@i2.org}
}
\maketitle
In this supplementary material, we apply our RdimKD on the language understanding, and provide the experimental details and some discussions for each experiment.
\section{Language understanding}
BERT\cite{devlin2018bert}, using language model pre-training trick, has significantly improved the performances of many Natural Language Processing tasks. However, the pre-trained language models are often too large and computationally intensive to deploy on resource-limited devices.

\textbf{TinyBERT: }To accelerate inference speed and reduce model size, the Noah's Ark Lab from Huawei proposes TinyBERT\cite{jiao2019tinybert} in which $\text{TinyBERT}_\text{6}$ with half transformer layers performs on-par with $\text{BERT}_\text{BASE}$ on GLUE\cite{wang2018glue} benchmark. They propose a novel Transformer distillation loss which composes of 3 parts: (1) Embedding loss from the input. (2) Attention + hidden loss from middle transformer layers. (3) Prediction loss from the output layer. Notice: the above hidden loss in step 2 is the same as no\_proj setting in our paper. And the whole training process contains 3 steps: (1) General distillation. (2) Data augmentation (3) Task-specific distillation.

In general distillation step, they use a large dataset to align the feature map distribution in the intermediate transformer layers which is time and resource consuming, so we just use the general distilled $\text{TinyBERT}_{\text{6}-\text{general}}$\footnote{https://huggingface.co/huawei-noah/TinyBERT\_General\_6L\_768D} pretrained model from huawei-noah's repo in huggingface. And we believe we will get a better result if we use our RdimKD from this general distillation step.

For task-specific distillation, the training process is also divided into 2 steps (each has a slightly different hyper-parameter setting). 
In step 1, they use the first two loss terms to train for several epochs depending on the size of each specific task dataset. 
In step 2, they use the last loss term to train for 3 epochs. 
We remove the fit\_dense layer for student model which exists even if the shape of hidden activations are the same for student and teacher in the original implementation and use our RdimKD methods to get the hidden loss term in step 1.
Our implementation code is largely based on the official TinyBERT repo\footnote{https://github.com/yinmingjun/TinyBERT} and we finetune the $\text{BERT}_\text{BASE}$ on our own to get the teacher models for our experiments.  We just use the original dataset instead of augmented dataset. 
 The results are shown in~\cref{tab:Natural language processing}. We can see that although the
$\text{TinyBERT}_{\text{6}-\text{general}}$ is pre-trained with no\_proj loss in general distillation step, our RdimKD can always beat the no\_proj method except SST-2. We also find out that when the task dataset is small, the experiment results can fluctuate a lot using no\_proj, which doesn't appear using RdimKD, benefited from the relaxed distillation.

\begin{table*}
\setlength\tabcolsep{6.5pt}
\begin{tabular}{l|c|c|c|c|c|c|c|c}
\toprule
& MNLI(m/mm)    & QQP    & QNLI    & SST-2    & CoLA(mcc)    & STS-B(corr)    & MRPC    & RTE\\
\hline
$\text{BERT}_\text{BASE}$(Teacher)  & 84.74/85.38    & 91.13    & 92.09    &93.81    & 60.93    & 89.00    & 87.99    &71.84    \\
\hline
no\_proj(TinyBERT) & 84.30/84.74    & 91.41    & 91.20    &\textbf{92.47}    & 50.29    & 89.01    & 85.29    &65.22    \\
RdimKD-R    & \textbf{84.49/84.77}    & 91.40    & 91.26    &92.28    & \textbf{51.28}    & \textbf{89.16}    & 86.76    &69.79    \\
RdimKD-A & 84.50/84.70    & 91.42    & \textbf{91.29}    &92.39    & 50.05    & 89.04    & \textbf{87.42}    &\textbf{69.92}    \\
RdimKD-P & 84.52/84.62    & \textbf{91.44}    & 91.25    &92.36    & 50.29    & 88.93    & 85.78    &68.23    \\
\bottomrule
\end{tabular}
\caption{Results on GLUE. We finetune $\text{BERT}_\text{BASE}$ as teacher models and use $\text{TinyBERT}_\text{6-general}$ as student. Although the $\text{TinyBERT}_\text{6-general}$ is pre-trained with no\_proj loss in general distillation step, our RdimKD can always beat the no\_proj method except SST-2. Notice: we use the same metric as TinyBERT\cite{jiao2019tinybert} (acc: QQP, QNLI, SST-2, MRPC, RTE, MNLI(m/mm: matched/mismatched dev set). mcc: CoLA. corr: STS-B). All of our results are the average on 3 trials with different random seeds.} 

\label{tab:Natural language processing}
\end{table*}

\section{Other experimental details}
\subsection{ImageNet classification}
The values of $\alpha$ and $r$ are shown in~\cref{tab:supp_cls}. The weight decay for MobileNet is 0.00001, and that for ResNet-18 is 0.00005. All the experiments are done on two A100-SXM-80GB GPUs.

\begin{table}
  \centering
  \setlength\tabcolsep{5pt}
\begin{tabular}{c|cc|cc}
\toprule
\multirow{2}{*}{} & \multicolumn{2}{c|}{MobileNet} & \multicolumn{2}{c}{ResNet18} \\
\cline{2-5}
                  & $\alpha$           & $r$           & $\alpha$          & $r$          \\
\hline
RdimKD-R          & 16               & 8           & 1              & 4          \\
RdimKD-A          & 8               & 4           & 1              & 4          \\
RdimKD-P          & 4               & 4           & 1              & 4         \\
\bottomrule
\end{tabular}
  \caption{The values of $\alpha$ and $r$ for image classification on ``ResNet-34 to ResNet-18'' and ``ResNet-50 to MobileNet'' in our experiments.}
  \label{tab:supp_cls}
\end{table}

\subsection{Object detection}
The values of $\alpha$ and $r$ for object detection are shown in~\cref{tab:supp_det}. For $r$, we initially set it to 4 for all, but later in the ablation experiment, we found that 2 is better for ``Faster-RCNN-FPN-ResNet-101 to Faster-RCNN-FPN-ResNet-18''. So, there may be better $r$ for other experiments. 

As for $\alpha$, we find that too large value may cause a training explosion. This is probably because we have fixed the batch normalization (BN). So we take the largest value that just doesn't cause explosion. It is interesting that the critical value of $\alpha$ for RdimKD-P is smaller than that for RdimKD-R. This is to be expected because when we project the feature maps onto the principal axes, the variance of the projected feature maps is much larger than when we project onto random axes. Another example is the comparison between RdimKD-P and pca\_last. The variance of samples after being projected to the first $d$ principal axes is much larger than after being projected to the last $d$ principal axes, the sum of the first $d$ eigenvalues is much larger than the sum of the last $d$ eigenvalues). And we find in experiments that the critical $\alpha$ for pca\_last is much larger than that for RdimKD-P.

The weight decay is 0.0001 for ResNet series, and 0.00004 for MobileNet\_V2. Other training details are the same as the standard protocols in the widely used Detectron2 library~\cite{wu2019detectron2}.

All the experiments are done on four A100-SXM-80GB GPUs.
\begin{table}
  \centering
  \setlength\tabcolsep{4pt}
\begin{tabular}{c|cc|cc|cc|cc|cc}
\toprule
\multirow{2}{*}{} & \multicolumn{2}{c|}{RR50}                   & \multicolumn{2}{c|}{XR50}                   & \multicolumn{2}{c|}{FR50} & \multicolumn{2}{c|}{FR18} & \multicolumn{2}{c}{FMV2} \\
\cline{2-11}
                  & \multicolumn{1}{c}{$\alpha$} & \multicolumn{1}{c|}{$r$} & \multicolumn{1}{c}{$\alpha$} & \multicolumn{1}{c|}{$r$} & $\alpha$            & $r$            & $\alpha$            & $r$            & $\alpha$            & $r$            \\
\hline
RdimKD-R          & 1                         & 4                     & 1                         & 4                     & 1                & 4            & 1                & 2            & 1                & 4            \\
RdimKD-A          & 1                         & 4                     & 1                         & 4                     & 0.3              & 4            & 0.4              & 2            & 0.5              & 4            \\
RdimKD-P          & 0.5                       & 4                     & 0.5                       & 4                     & 0.15             & 4            & 0.2              & 2            & 0.3              & 4           \\
\bottomrule
\end{tabular}
  \caption{The values of $\alpha$ and $r$ for object detection in our experiments. RR50 is ``RetinaNet-ResNet-101 to RetinaNet-ResNet-50'', B is ``RetinaNet-ResNeXt-101 to RetinaNet-ResNet-50'', FR50 is ``Faster-RCNN-FPN-ResNet-101 to Faster-RCNN-FPN-ResNet-50'', FR18 is ``Faster-RCNN-FPN-ResNet-101 to Faster-RCNN-FPN-ResNet-18'', and FMV2 is ``Faster-RCNN-FPN-ResNet-101 to Faster-RCNN-FPN-MobileNet V2''. }
  \label{tab:supp_det}
\end{table}

\subsection{Semantic segmentation}
The values of $\alpha$ and $r$ for segmentation are shown in~\cref{tab:supp_seg}. The weight decay is 0.0001 for both the two experiments. We find that the fluctuations in the results of each experiment is relatively large, so all the results are the average of six trials.
All the experiments are done on one A100-SXM-80GB GPUs.

\begin{table}
  \centering
  \setlength\tabcolsep{5pt}
\begin{tabular}{c|cc|cc}
\toprule
\multirow{2}{*}{} & \multicolumn{2}{c|}{MobileNet\_V2} & \multicolumn{2}{c}{ResNet18} \\
\cline{2-5}
                  & $\alpha$           & $r$           & $\alpha$          & $r$          \\
\hline
RdimKD-R          & 4               & 8           & 4              & 8          \\
RdimKD-A          & 4               & 8           & 4              & 8          \\
RdimKD-P          & 4               & 8           & 2              & 8         \\
\bottomrule
\end{tabular}
  \caption{The values of $\alpha$ and $r$ for semantic segmentation on ``DeepLabv3+-ResNet-101 to DeepLabv3+-MobileNet\_V2'' and ``DeepLabv3+-ResNet-101 to DeepLabv3+-ResNet-18'' in our experiments.}
  \label{tab:supp_seg}
\end{table}

\subsection{Speech recognition}
In this experiment, each trial is run over 40+ hours on eight A100-SXM-80GB GPUs. Due to the high cost, each result is from only one trial. $\{\alpha= 2, r= 4\}$. We trained 50 epochs very trial. Other training details are same to the Wenet Library~\cite{yao2021wenet, zhang2022wenet}.


{\small
\bibliographystyle{ieee_fullname}
\bibliography{egbib}
}